# Order to Disorder Transitions in Hybrid Intelligent Systems: a Hatch to the Interactions of Nations -Governments

Hamed Owladeghaffari[*]

[*]Department of Mining and Metallurgical Engineering, Amirkabir University of Technology, Tehran, Iran -h.o.ghaffari@gmail.com

## Abstract

*In this study, under general frame of MAny Connected Intelligent Particles Systems (MACIPS), we reproduce two new simple subsets of such intelligent complex network, namely hybrid intelligent systems, involved a few prominent intelligent computing and approximate reasoning methods: self organizing feature map (SOM), Neuro-Fuzzy Inference System and Rough Set Theory (RST). Over this, we show how our algorithms can be construed as a linkage of government-society interaction, where government catches various fashions of behavior: "solid (absolute) or flexible". So, transition of such society, by changing of connectivity parameters (noise) from order to disorder is inferred. Add to this, one may find an indirect mapping among finical systems and eventual market fluctuations with MACIPS.*

## 1. Introduction

Complex systems are often congruous with uncertainty and order-disorder transitions. Apart of uncertainty, fluctuations forces due to competition of between constructive particles of system drive the system towards order and disorder. There are prominent examples which their behaviors show such anomalies in their evolution, i.e., physical systems, biological, financial systems [1]. In other view, in monitoring of most complex systems, there are some generic challenges for example sparse essence, conflicts in different levels, inaccuracy and limitation of measurements ,which in beyond of inherent feature of such interacted systems are real obstacle in their analysis and predicating of behaviors. There are many methods to analyzing of systems include many particles that are acting on each other, for example statistical methods [2], Vicsek model [3]. Other solution is finding out of "main nominations of each distinct behavior which may has overlapping, in part, to others". This advance is to bate of some mentioned difficulties that can be concluded in the "information granules" proposed by Zadeh [4]. In fact, more complex systems in their natural shape can be described in the sense of networks, which are made of connections among the units. These units are several facets of information granules as well as clusters, groups, communities, modules [5]. Let us consider a more real feature: dynamic natural particles in their inherent properties have (had have-will have) several appearances of "natural" attributes as in individually or in group forms. On the other hand, in society, interacting of main such characteristics (or may extra-natural forces: metaphysic) in facing of predictable or unpredictable events, determines destination of the supposed society.

Based upon the above, hierarchical nature of complex systems [6], developed (developing) several branches of natural computing (and related limbs) [7], collaborations, conflicts, emotions and other features of real complex systems, we propose a general framework of the known computing methods in the connected (or complex hybrid) shape, so that the aim is to inferring of the substantial behaviors of intricate and entangled large societies. Obviously, connections between units of computing cores (intelligent particles) can introduce part (or may full) of the compartments (demeanors-deportments...). Complexity of this system, called MAny Connected Intelligent Particles Systems (MACIPS), add to reactions of particles against information flow, can open new horizons in studying of this big query: *is there a unified theory for the ways in which elements of a system(or aggregation of systems) organize themselves to produce a behavior*?[8]. With expanding of a few MACIPS (fig.1.) within a network, we may construe events of our world within small world. Considering of growing, evolution, cliquing, competition and collaboration among supposed networks [9] can instill a concomitant strategy on the insatiable problems of our world (fig.2.).

In this study, we select a very little limited part of MACIPS, as well as hybrid intelligent systems: Self Organizing Neruo-Fuzzy Inference System (SONFIS) and Self Organizing Rough Set Theory (SORST), and then investigate several levels of responses in facing with the real information. We show how relatively such our simple methods that can produce (mimic) complicated behavior of government-nation interactions .Mutual relations between proposed algorithms layers identify order-disorder transferring of such systems. Developing of such intelligent hierarchical networks, investigations of their performances on the noisy information and exploration of possible relate between phase transition steps of the MACIPS and flow of information in to such systems are new interesting fields, as well in various fields of science and economy.

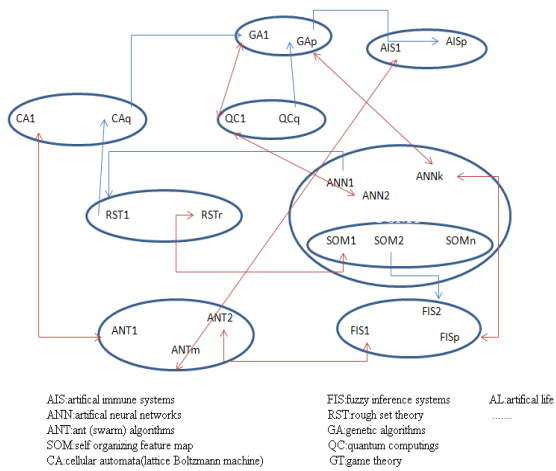

**Figure 1. A schematic view of MAny Connected Intelligent Particles Systems (MACIPS)**

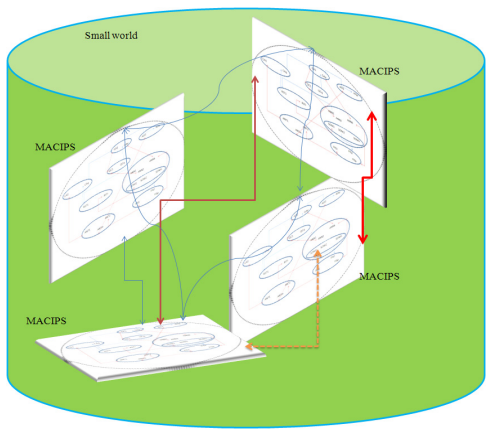

**Figure 2. A schematic Small world perspective: interactions of some MACIPS within network**

## 2. Employed methods

In this section based upon Self Organizing feature Map (SOM) [10], adaptive Neuro-Fuzzy Inference System (NFIS) [11] and Rough Set Theory (RST) [12], [13], we reproduce a very little part of MACIPS: Self Organizing Neuro-Fuzzy Inference System (SONFIS) and Self Organizing Rough Set Theory (SORST). Indeed we have developed dependency rule generation –RST- in MatLab7, and on this added toolbox other appropriate algorithms have been prepared [14], [15], and [16]. In this study our aim is to investigate of order-disorder transition in the mentioned systems.

### 2.1. SONFIS & SORST

Developed algorithms use four basic axioms upon the balancing of the successive granules assumption:

- Step (1): dividing the monitored data into groups of training and testing data
- Step (2): first granulation (crisp) by SOM or other crisp granulation methods
  Step (2-1): selecting the level of granularity randomly or depend on the obtained error from the NFIS or RST (regular neuron growth)
  Step (2-2): construction of the granules (crisp).
- Step (3): second granulation (fuzzy or rough granules) by NFIS or RST
  Step (3-1): crisp granules as a new data.
  Step (3-2): selecting the level of granularity; (Error level, number of rules, strength threshold, scaling of inserted information...)
  Step (3-3): checking the suitability. (Close-open iteration: referring to the real data and reinspect closed world)
  Step (3-4): construction of fuzzy/rough granules.
- Step (4): extraction of knowledge rules

Selection of initial crisp granules can be supposed as "Close World Assumption (CWA)" .But in many applications, the assumption of complete information is not feasible, and only cannot be used. In such cases, an "Open World Assumption (OWA)', where information not known by an agent is assumed to be unknown, is often accepted [17].

Balancing assumption is satisfied by the close-open iterations: this process is a guideline to balancing of crisp and sub fuzzy/rough granules by some random/regular selection of initial granules or other optimal structures and increment of supporting rules (fuzzy partitions or increasing of lower /upper approximations ), gradually. The overall schematic of SONFIS and SORST-AS has been shown in Fig.3, Fig.4. In first granulation step, we use a linear relation is given by:

$$N_{t+1} = \alpha N_t + \Delta_t ; \Delta_t = \beta E_t + \gamma \qquad (1)$$

Where $N_t = n_1 \times n_2; |n_1 - n_2| = Min.$ is number of neurons in SOM or Neuron Growth (NG); $E_t$ is the obtained error (measured error) from second granulation on the test data and coefficients must be determined, depend on the used data set. Granulation level is controlled with four main parameters: range of neuron growth, number of rules, number of discretization of attributes in RST and/or error level. The main benefit of SONFIS is to looking for best structure and rules for two known intelligent system, while in independent situations each of them has some appropriate problems such: finding of spurious patterns for the large data sets, extra-time training of NFIS or SOM.

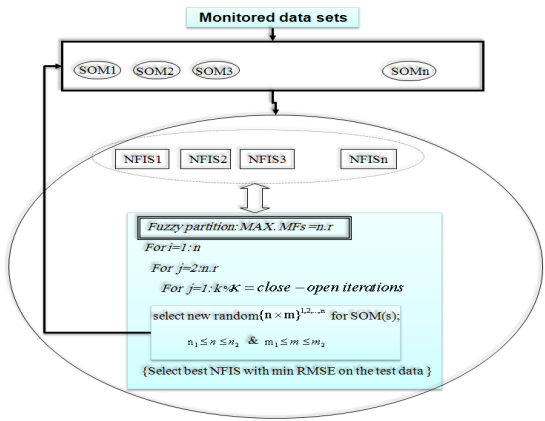

**Figure 3. Self Organizing Neuro-Fuzzy Inference System (SONFIS)**

Applying of SOM as a preprocessing step and discretization tool is second process, in SORST. Categorization of attributes (inputs/outputs) is transferring of the attribute space to the symbolic appropriate attributes. In fact for continuous valued attributes, the feature space needs to be discretized for defining indiscernibilty relations and equivalence classes. We show how the changing of scaling induces a phase transition step, while first granulation level is updated by a similar linear relation. Because of the generated rules by a rough set are coarse and therefore need to be fine-tuned, here, we have used the preprocessing step on data set to crisp granulation by SOM (close world assumption). In fact, with referring to the instinct of the human, we understand that human being want to states the events in the best simple words, sentences, rules, functions and so forth. Undoubtedly, such granules while satisfies the mentioned axiom that describe the distinguished initial structure(s) of events or immature data sets. Second SOM, as well as close world assumption, gets such dominant structures on the real data. In other paper [17], we have shown under a constant scaling value, SORST redisplays a transferring step when noise parameters ($\alpha$, $\beta$ and $\gamma$) in relation1 are turned.

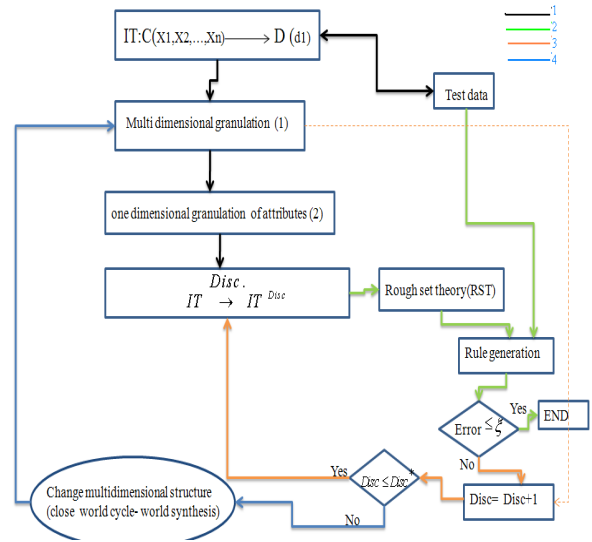

**Figure 4. Self Organizing Rough Set Theory-Adaptive Scaling (SORST-AS)**

Apart of the aforesaid background behind the proposed algorithms, we can assume interactions of the two layers of algorithms as the behaviors of complex systems such: society and government, where reactions of a dynamic community to an "absolute (solid-dictatorship) or flexible (democratic)" government (regulator) is controlled by correlation (noise) factors of the two simplified systems. In absolute case, the second layer (government / regulator) has limited rules with stable learning iteration for all of matters. In first layer, society selects most main structures of the stimulator where these clusters upon the reaction of government and pervious picked out structures will be adjusted. In facing this, flexible regulator has ability of adapting with the evolution of society.

This situation can be covered by two discrete alternatives: evolution of constitutive rules (policies) over time passing or a general approximation of the dominant rules on the emerged attitudes. In latter case the legislators can considers being conflicts of the emerged states. Other mode can be imagined as poor-revealing structures of the society due to poor-learning or relatively high disturbances within inner layers of the community. It must be noticed, we may choose other two general connected networks or other natural inspired systems involve such hierarchical topology for instances: stock market and stock holders, queen and bees, confliction and quarrel between two countries, interaction among nations (so its outcome can be strategy identifying for trade barriers[20]) and so on. With variations of correlation factors of the two sides (or more connected intelligent particles), one can identify point (or interval) of changes behavior of society or overall system and then controlling of society may be satisfied. The scaling of the observed information, definitely, has a great role in identification of long and short politics for especially democratic government. In this paper we show how with changing of scaling, democratic government may lose the current prevalent order of the society while noise parameters keep on the initial values. In [16], we have evaluated the variations of noise parameters in SORST. Obviously, considering of simple and reasonable phase transition measure in the mentioned systems will be necessary so that regarding of entropy and order parameter are two distinguished criteria [1].

In this study we use crisp granules level to emulation of phase passing. So, we consider both "absolute and flexible" government while in latter case the approximated rules are contemplated.

## 3. Phase transitions: how does a society go into the revolutionary state?

In this part of paper, we ensue our algorithms on the "lugeon data set" [14]. This study only considers phase transition view of our proposed algorithms and direct applications of the mentioned systems in other data sets can be found in [14], [15]. To evaluate the interactions due to the lugeon values we follow two situations where phase transition measure is upon the crisp granules (here NG): 1) second layer takes a few limited rules by using NFIS; 2) second layer keep all of extracted rules by RST and under an approximated progressing (with changing of scaling).

Analysis of first situation is started off by setting number of close-open iteration and maximum number of rules equal to 30 and 3 in SONFIS respectively. The error measure criterion in SONFIS is Root Mean Square Error (RMSE), given as below:

$$RMSE = \sqrt{\frac{\sum_{i=1}^{m}(t_i - t_i^*)^2}{m}}\;;$$

where $t_i$ is output of SONFIS and $t_i^*$ is real answer; $m$ is the number of test data (test objects). In the rest of paper, let $m=93$ and number of inserting data set $=600$. By employing of (1) in SONFIS and $\beta=.001$ and $\gamma=.5$; the general patterns of NG and RMSE vs. time steps and variations of $\alpha$ can be observed (fig.5, 6). It must be noticed for two like process (i.e., $\alpha=.9$), we may have different situation of neuron growth.

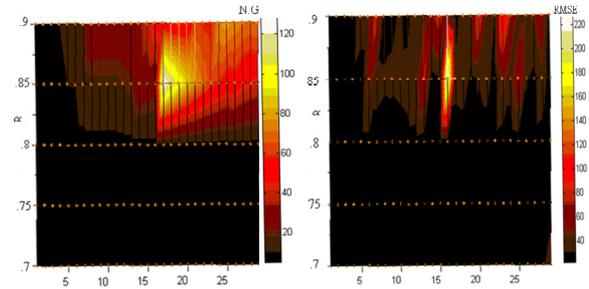

**Figure 5.** Effect of Alpha variations in neuron growth –Beta=.001(N.G)-left- &RMSE-right- of SONFIS with n.r=2 over 30 iterations; .7=< Alpha<=.9

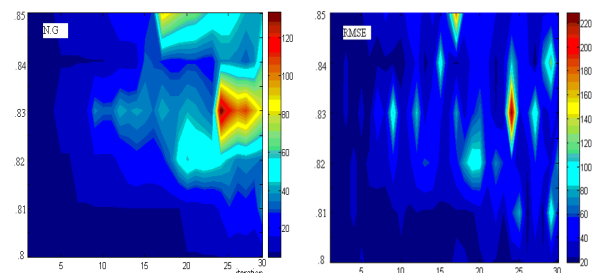

**Figure 6.** Micro-view on the Alpha variations in neuron growth –Beta=.001(N.G)-left- &RMSE-right- of SONFIS with n.r=2 over 30 iterations; .8=< Alpha<=.85

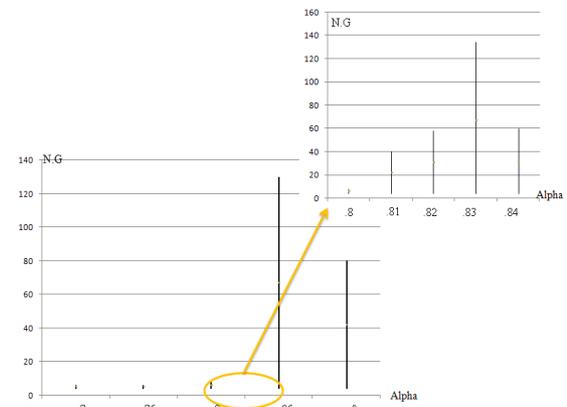

**Figure 7.** Aggregation of Alpha variations N.G in SONFIS with n.r=2 over 30 iterations (Phase transition step)

The main reason of such behavior is on the regulation of weight neurons in SOM thank to initial random selection and fall in to the "dead neurons state". However, this will be interesting if we see real case, as is appeared in real society, in order to "in an identical cases (but in an unlike iteration) society may shows other behavior, not completely different from other mate". Fig.5 indicates how the neurons fluctuations with time passing reveal more chaos while the phase transition step can be transpired in $\alpha=.8-.85$. This evolution of first layer has occurred in a continuous way, as well as other progressing of swarms systems [18]. The integration of average neuron growth, in a bar graph, exhibits phase transferring interval (fig.7). We can reap other interesting results when the numbers of rules are added while the government persists on the same number of learning (i.e., absolute government) [16].

Let us consider a reverse occasion: $\alpha$ is constant (=.9) and $\beta$ takes different values (fig.8). Such consideration, apart of distinction of the possible phase altering step (after $\beta=4\times10^{-4}$), may display another feature of society alteration: the proper chaos related to the later fashion has larger values so that is not relatively agreed with N.G. In fact, our government loses pervious relative order. In both two former and latter options, the phase transition has been occurred gradationally likewise one can consider three discrete steps to these conversions: society with "silent dead (laminar)", in transition and in triggering of revolutionary community. Now, the possible question may come in to the mind how the sensitivity of first layer to those parameters is. To answer this query, in other process, simultaneously $\alpha$ and $\beta$ are changed over 10 iterations (fig.9.). In a scaled coordination, it can be probed that the rate of $\alpha/\beta$ for the lugeon data set and for laminar state is near to 1, even if $\alpha$ takes a little bigger value. This feature expresses response of overall system to the current wave of data, is more depends on the pervious state of society than to the government reaction. Another interesting result of such accomplishing is that the behavior of the system in large values of $\alpha$ and $\beta$: in despite of some anomalies, for large digits system doesn't consider other disorder parameter and fall in to the disordering way.

In second situation we employ SORST-AS, upon this assumption that the government based on history, experience and other like fashions in the world, has ability to elicitation of relatively approximated rules of the observed and distinguished behaviors (by transferring of attributes to the changeable scaled classes using 1-D SOM, as well as low, middle, high and so on). The applied error measure for measure of performance of RST is given by:

$$MSE = \frac{\sum_{i=1}^{m}(d_i^{real}-d_i^{classified})^2}{m};$$

In deducing of decision for approximated rules (not unique decision part), we select highest value (largest ambiguities) for such decisions. By repeating of steps as well pervious situation, we obtain behavior of SORST-AS where we employ $\alpha=.9$, $\beta=.7$ and $\gamma=1$, in equation 1(fig.10). This proves how with changing of scaling, democratic government may lose the current prevalent order of the society while noise parameters keep on the initial values. In other word, more details may astound even democratic government. From rough set theory view, finding best bins rescues the poor performance of algorithm and preserves suitable (highest) dependency.

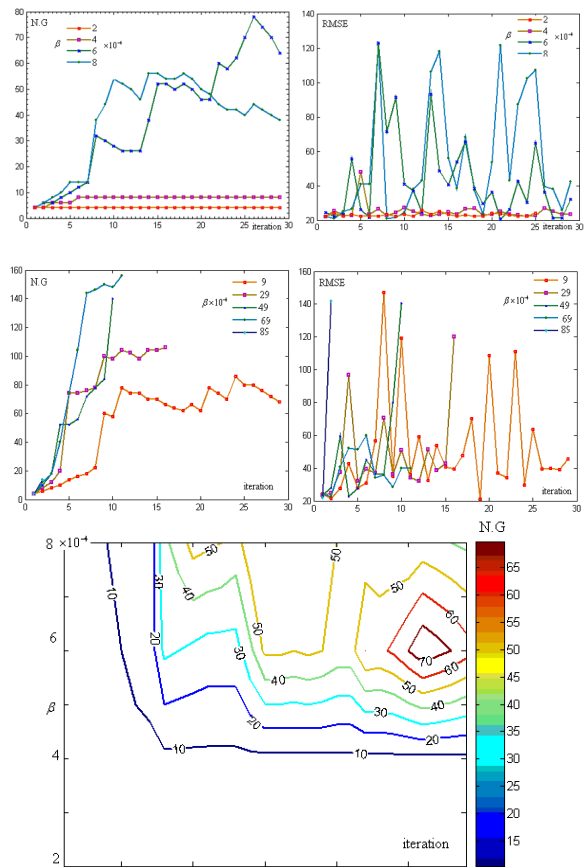

**Figure 8. Effect of Beta variations-Alpha =.9 - in Neuron Growth (N.G)-left- &RMSE-right- of SONFIS with n.r=2 over 30 iterations; a) 2*10^-4=< Beta<=8*10^-4; b) 9*10^-4=< Beta<=85*10^-4 and c) Phase transition diagram contour of N.G-SONFIS**

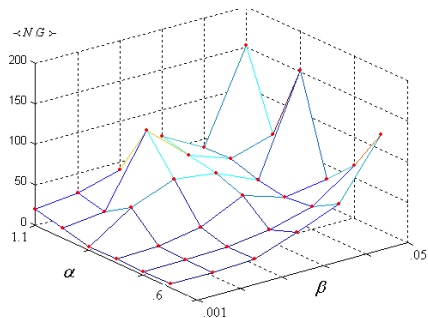

**Figure 9. Effect of Alpha & Beta variations over 10 iterations on the average NG-n.r=2 (SONFIS)**

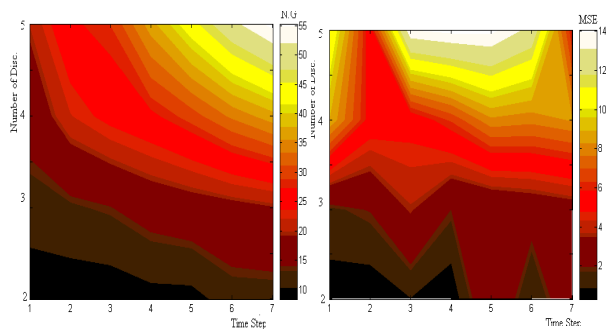

**Figure 10. Color code of Disc. Numbers variations in SORST on the N.G and MSE over 7 time steps-($\alpha$ =.9, $\beta$ =.7 and $\gamma$ =1)**

We can investigate other more complex cases, for example in [19]; we have explained how the rate of scaling of attributes displays other phase transition state. One may investigate the being of elate among entropy of actual data set and transitions points.
Answering to similar questions can instill crucial ideas in complex systems, i.e., how real disturb wave can be controlled by government (or other regulator), is will be necessary to modifying of disorders factors, what will be the extra-forces ($\gamma$) influences.

## 4. Conclusion

In this study we proposed two new algorithms in which SOM, NFIS and RST, based on general frame of MAny Connected Intelligent Particles Systems (MACIPS), make SONFIS and SORST. Main idea behind our algorithms is to finding out of best reduced objects, are in balance with second granulation level. Mutual relations between algorithms layers identify order-disorder transferring of such systems. So, we found our proposed methods have good ability in mimicking of government-nation interactions while government and society can take the different states of responses. Developing of such intelligent hierarchical networks, investigations of their performances on the noisy information and exploration of possible relate between phase transition steps of the MACIPS and flow of information in to such systems are new interesting fields, as well in various fields of science and economy.